\documentclass[pdflatex,sn-mathphys-num]{sn-jnl}

\usepackage{graphicx}%
\usepackage{multirow}%
\usepackage{amsmath,amssymb,amsfonts}%
\usepackage{amsthm}%
\usepackage{mathrsfs}%
\usepackage[title]{appendix}%
\usepackage{xcolor}%
\usepackage{textcomp}%
\usepackage{manyfoot}%
\usepackage{booktabs}%
\usepackage{algorithm}%
\usepackage{algorithmicx}%
\usepackage{algpseudocode}%
\usepackage{listings}%

\theoremstyle{thmstyleone}%

\theoremstyle{thmstyletwo}%

\theoremstyle{thmstylethree}%

\begin{document}

\title[Temperament Engineering]{Temperament Engineering: Designing Strategic Behavioural Diversity in Robot Swarms}

\author*[1]{\fnm{Edmund R.} \sur{Hunt}}\email{edmund.hunt@bristol.ac.uk}

\affil*[1]{\orgdiv{School of Engineering Mathematics \& Technology}, \orgname{University of Bristol}, \orgaddress{\street{Ada Lovelace Building, Tankard's Cl}, \city{Bristol}, \postcode{BS8 1TW}, \country{United Kingdom}}}

\abstract{Behavioural heterogeneity is increasingly recognised as enabling robot swarm function, but lacks systematic design methods. In animal behaviour, `temperament' denotes the consistent individual differences in behaviour that persist across time and context; I propose `temperament engineering', a bio-inspired framework that treats the swarm's distribution of temperaments, rather than the individual controller, as the design object. Drawing on a finite set of evolutionarily validated behavioural axes, it calibrates the swarm's emergent adaptation to the mission ecology: risk against reward, novelty against familiarity, exertion against conservation.}


\maketitle
No two robots are truly identical. Even with best-efforts manufacturing and calibration, `quasi-homogeneity' \cite{Beni2005,hamann2018} is the best a swarm can attain, and the residual variation is usually treated as an imperfection to be minimised. In animal collectives the reverse holds: individual behavioural variation is shaped by evolution and often decisive for group performance \cite{Reale2007,BastilleRousseau2019,OShea-Wheller2021}. Such heterogeneity is increasingly recognised as an enabler for swarm function --- through uncalibrated `noise' \cite{Raoufi2023}, role specialisation \cite{Dorigo2020} or decision-rule diversity \cite{Zakir2024} --- yet the field lacks a systematic way to engineer the behavioural variation most relevant to field deployment. Calibration, battery state, sensor drift and manufacturing tolerance inevitably produce variation in how robots sense and act, with distributions typically approximately normal, reflecting their origin in many small independent physical sources \cite{Raoufi2023}. Because sensing and actuation feed directly into behaviour, this physical variation manifests as behavioural variation --- most directly on the activity axis, where motor characteristics and battery state set how fast and how far a robot moves, and indirectly on others, as when miscalibrated perception shifts the effective threshold at which a robot accepts risk. Along such axes the swarm already occupies a distribution of values rather than a single point. This is \textit{incidental} temperament variation, the unintended distribution produced by embodied robot variation. Incidental variation does not populate every behavioural axis alike: the axes governing interaction between agents (such as sociability) are less exposed to physical happenstance, so variation there must be introduced by design rather than inherited from hardware. Across all axes alike --- those where physical variation already impinges and those where it does not --- I argue this variation should be deliberately shaped, not minimised or overlooked, and actively engineered for the mission ecology: the interaction of task demands, environmental conditions and operational hazards that determines which behavioural trade-offs a deployment turns on.

\begin{figure}
\includegraphics[width=\textwidth]{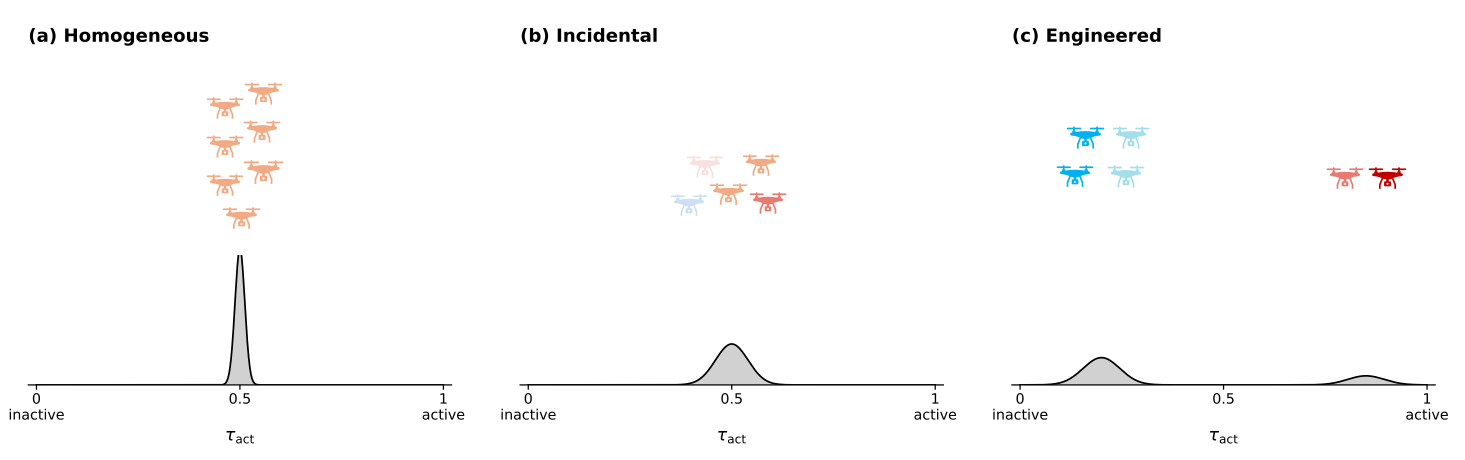}
\caption{\textbf{(Quasi-)homogeneous, incidental, and engineered temperament distributions along the activity-level axis}. Each panel shows a representative swarm of UAVs whose colour encodes $\tau_{\text{act}} \in [0,1]$, from blue (inactive) to red (active), above the underlying distribution. (a) A (quasi-) homogeneous swarm with all robots calibrated at $\tau_{\text{act}} = 0.5$. (b) Incidental temperament: an approximately normal distribution arising from factors such as battery state, motor characteristics, lack of calibration and mechanical wear that the designer has not addressed. (c) Engineered temperament: a deliberately designed bimodal distribution comprising a sustained-patrol majority and a small burst-response minority, designed for a mission requiring both sustained coverage and rapid reaction. The comprehensive design process considers the full behavioural hypervolume across multiple relevant axes.}
\label{fig1}
\end{figure}

\begin{figure}
\includegraphics[width=\textwidth]{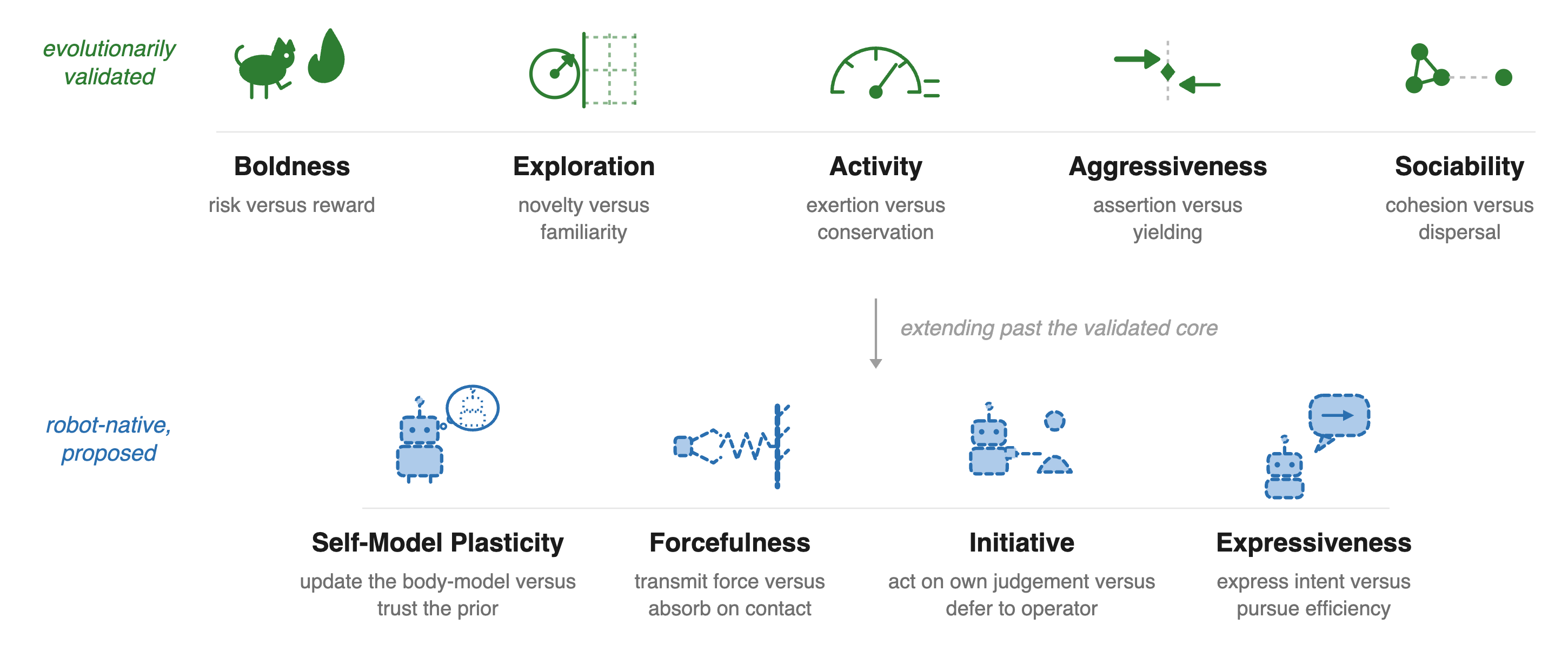}
\caption{\textbf{The evolutionarily validated temperament axes (top) and some tentative robot-native axes (bottom)}. The five foundational axes are evolutionarily validated in animal collectives and proposed here as a design vocabulary for robot swarms. Three respects in which robots differ from animals open further, robot-native dimensions. First, a robot's body is designed and explicitly modelled, so the model can be wrong and revised: this points toward \textit{Self-Model Plasticity}, a readiness to update the internal dynamics-model rather than hold to the prior. Second, a robot's body is not self-maintaining, and may absorb impact, reconfigure or be expended: this points toward \textit{Forcefulness}, trading precision and force transmission against energy absorption and safety on contact --- distinct from aggressiveness, which governs competition for resources rather than physical contact. Third, a robot operates under human supervision, which opens two distinct dimensions: \textit{Initiative}, how far a robot acts on its own judgement rather than deferring to the operator; and \textit{Expressiveness}, how far it shapes its behaviour to express intent and be legible to teammates and overseers, rather than pursuing raw efficiency. These robot-native axes are preliminary and non-exhaustive.}
\label{fig2}
\end{figure}

In animal behaviour research, `temperament' (or `personality') describes consistent individual differences in behaviour across time and contexts \cite{Reale2007}. Although `personality' is more common, and this nomenclature actively debated \cite{Beekman2017,Dingemanse2017,Bell2017,Briffa2017}, I follow R\'eale et al. \cite{Reale2007} in preferring `temperament', which softens inferences about psychological disposition --- doubly apt in robotics, where `personality' risks connotations of intent or sentience. Their widely adopted framework distinguishes five axes --- shyness–boldness, exploration–avoidance, activity level, aggressiveness and sociability \cite{Reale2007} --- which I consider below. Temperament is not task-level specialisation, which narrows an agent's repertoire to a role; any embodied agent already sits somewhere on each trade-off axis, whether or not the designer has chosen where. The designer's choice is therefore not whether to give a robot a temperament, but whether to design it deliberately or leave it an unintended consequence of controller and morphology (Figure~\ref{fig1}). Rather than reinvent a vocabulary, I import a finite set of trade-off axes that natural selection has already validated (Figure~\ref{fig2}, top) and recast it as an engineering design vocabulary; in the Outlook I tentatively suggest some axes unique to robotics (Figure~\ref{fig2}, bottom).

Field robotics faces many of the challenges that shaped temperament in animals, and in embodied intelligence adaptive behaviour arises from the interaction of control, morphology and environment \cite{Rus2015,Li2025EMAS}. A temperament's substrate need not lie entirely in software: a compliant, energy-absorbing body \cite{Rus2015,Pfeifer2006} inhabits a cost landscape favouring shy-typical behaviour, while a rigid, armoured platform shifts it toward boldness, with the integrity to act on risks a fragile platform could not survive. The coupling is loose: in birds, morphological--physiological integration is tight but body--behaviour coupling heterogeneous \cite{GaonaGordillo2023}, and engineers may recombine morphology and behaviour more freely than evolution does. Temperament is thus the behavioural counterpart to morphological design, and in heterogeneous platforms (e.g., \cite{Ducatelle2011,Thenius2016}) it is natural to align the two: a large, well-sensorised robot is costly to lose, so a cautious profile protects the investment, while a small, expendable one is the natural candidate for boldness. Morphology and temperament then jointly define a swarm's functional diversity.

This split in substrate has a consequence for how temperament can be assigned. The embodied component --- resident in morphology, calibration and mechanical state --- is fixed at the moment of action: it can be designed in advance but not re-bodied on demand. The software-resident component, by contrast, is a control parameter that can in principle be set, or reset, at deployment time or later. Temperament is therefore only partly a property a robot carries; in its software-resident part it is also a distribution that can be re-derived when circumstances change, conditioned on which robots are present and where --- a flexibility unavailable to organisms, whose temperament is wholly embodied.

Whether fixed or reassignable, what matters is the distribution: a swarm whose robots share an identical behavioural strategy may be well adapted to a narrow range of conditions but brittle when conditions vary. Mechanisms such as online response-threshold adaptation \cite{Bonabeau1997,Castello2016} can introduce post-deployment heterogeneity within an otherwise homogeneous controller, but this variation is restricted to the parameters the controller exposes; temperament engineering operates above this level, treating the strategic axes of variation themselves as design variables. The behavioural diversity of biological collectives can be understood as an anticipatory adaptation to their ecology --- and the swarm's behavioural distribution is the property temperament engineering sets out to design deliberately.

\section*{The Temperament Engineering Workflow}

Temperament engineering turns these insights into a structured design method for robot swarms. I identify three design activities --- Trait Mapping, Distribution Planning, and Plasticity Tuning (how far, if at all, a robot adjusts its temperament as its circumstances change) --- which together furnish an informed starting point for the temperament parameters and reaction norms of a deployed swarm (Figure~\ref{fig3}, Panel 1). This phase is top-down: it fixes the relevant axes and supplies a prior on how their values might be distributed, narrowing the search space to something tractable. The verified distribution and reaction-norm parameters that emerge from Phase 2 are guided by this prior but not constrained to it --- bottom-up refinement may steer the design away from the initial proposal --- before deployment, where they interact with the mission ecology to produce emergent collective behaviour (Panel 3).

\subsection*{From mission to temperament}

\begin{figure}
\includegraphics[width=\textwidth]{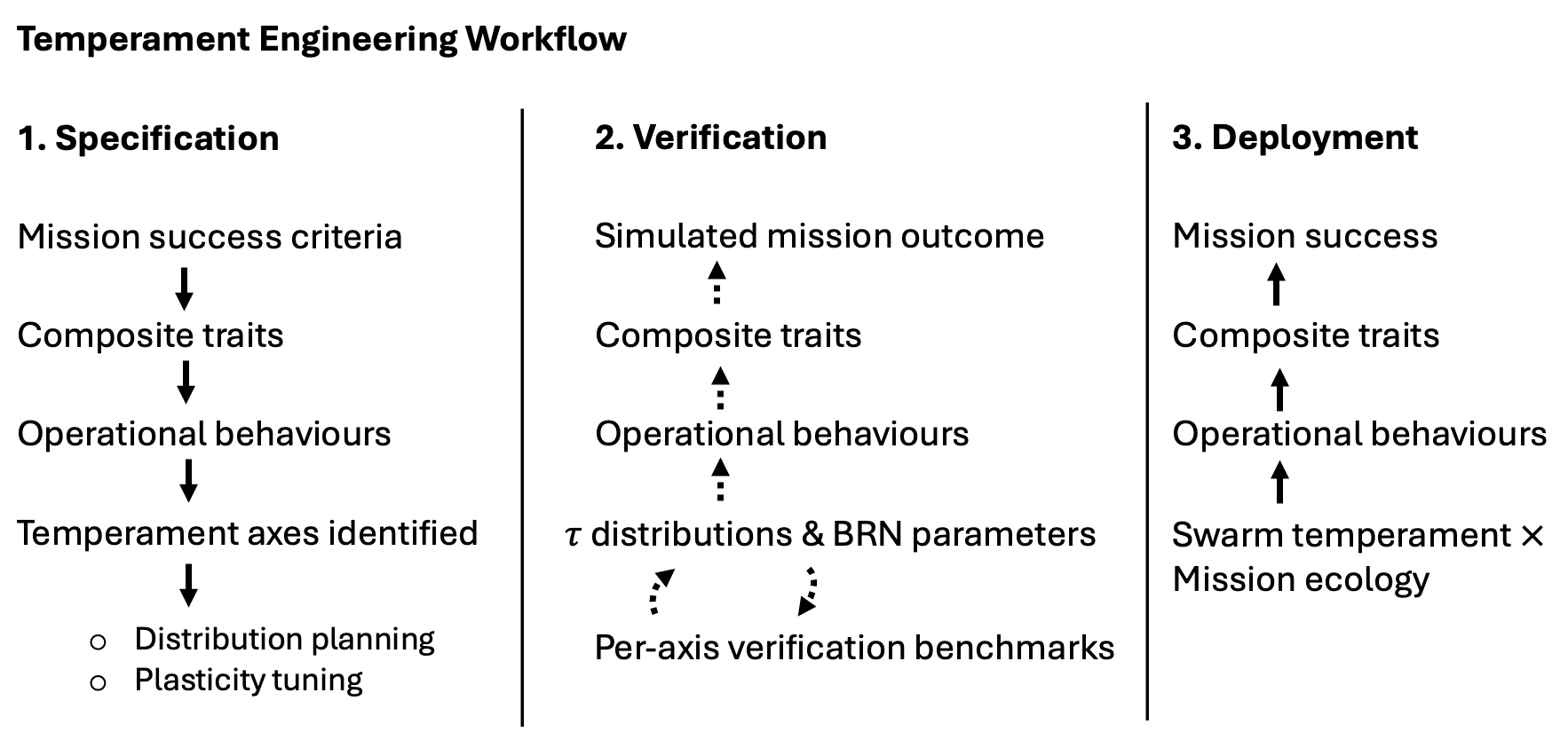}
\caption{The temperament engineering workflow comprises three phases. (1) \textbf{Specification (top-down):} mission success criteria are decomposed downward through required composite traits and operational behaviours into an identified set of relevant temperament axes (trait mapping). The designer then proposes how the trait(s) are distributed across the swarm (distribution planning) and whether robots adjust their temperament in response to environmental cues via behavioural reaction norms (plasticity tuning, optional), furnishing a prior on the design. (2) \textbf{Verification:} this proposal seeds a bottom-up search --- the proposed $\tau$ distributions and BRN parameters are refined and validated through increasingly integrated test scenarios, from per-axis calibration up to whole-mission rehearsal, where the realised spatial structure is checked against intent.The bottom loop indicates iterative parameter refinement, which may revise the proposal. (3) \textbf{Deployment:} the verified temperament parameters interact with the actual mission ecology to produce operational behaviours that determine mission success. The solid arrow style indicates the top-down prior (panel 1) and causal emergence (panel 3); the dashed, bottom-up refinement and verification (panel 2). The hierarchical structure is modelled after Réale et al. \cite{Reale2007}.}
\label{fig3}
\end{figure}

Trait Mapping is deliberately general: any deployment descends the same hierarchy --- from mission success criteria, to composite traits, to operational behaviours, to the temperament axes that shape them (Figure~\ref{fig3}). This inverts the hierarchy R\'eale et al.\ \cite{Reale2007} use to relate temperament to biological fitness through behaviour, with mission success as the engineering analogue of fitness. For instance, `comprehensive contamination mapping with minimal robot loss' yields composite traits `thorough coverage' and `survivability', whose operational behaviours --- `enter elevated-radiation zones', `investigate novel sensor anomalies' --- map onto the shyness--boldness and exploration--avoidance axes. The descent is rarely a clean tree: one composite trait may draw on several axes, and one axis may serve several traits, so identifying the right axes for a mission is a matter of judgement rather than mechanical decomposition. That judgement --- anticipating which behavioural trade-offs a deployment will actually turn on --- is where domain expertise enters the workflow, supplying the insightful prior the bottom-up search will later refine. Reasoning in temperament rather than bespoke task-specific parameters gives these system-level trade-offs a transferable, biologically grounded vocabulary that serves across very different deployments.

The remainder of this section develops the specification phase, which operates throughout on a single primitive: the temperament parameter $\tau$ and its reaction norm. Trait Mapping, above, identifies the relevant axes; Distribution Planning and Plasticity Tuning then shape $\tau$ across the swarm. Both presuppose the primitive on which they act, so I first make it precise --- how each axis is rendered as a control parameter (\emph{Defining the primitive}, below) --- before turning to the two activities that distribute and tune it (\emph{Shaping the distribution}).

\subsection*{Defining the primitive}

\paragraph*{Per-axis operationalisation}

I propose that each temperament axis corresponds to a continuous modifier $\tau \in [0,1]$ on the robot's decision-making process. Each axis names a behavioural trade-off that field robots must resolve and that the behavioural-ecology literature has already characterised and validated; for each, I give the trade-off it captures and a representative control-theoretic operationalisation. The correspondences are illustrative rather than definitive.

\begin{itemize}
\item \textbf{Boldness} governs risk appetite: the willingness to act in the face of known hazard, such as entering zones of elevated danger. This is the domain of risk-aware planning and safety-critical control, where the question is not whether to avoid risk but how much to accept for a prospective reward \cite{Majumdar2020,Ames2017}. A scalar weight $\tau_{\text{bold}}$ trades known risk against potential reward. Following R\'eale et al.\ \cite{Reale2007}, this axis concerns response to \emph{known} risk rather than novelty, which is captured separately by exploration--avoidance.

\item \textbf{Exploration} governs novelty-seeking: investigating unfamiliar or unmapped regions versus exploiting the well-characterised. This is the exploration--exploitation trade-off in autonomous mapping and active perception, where information-theoretic controllers weigh expected information gain against the cost of leaving known, productive areas \cite{Julian2014,McGuire2019}. A parameter $\tau_{\text{exp}}$ shifts an objective function from task-driven cost minimisation towards information gain.

\item \textbf{Activity} governs exertion and its costs: the intensity of behaviour, from sustained low-tempo patrol to burst-mode rapid response, and the maintenance of a reserve workforce or distributed energy store \cite{Charbonneau2017,Melhuish2007}. Beyond energy, high activity hastens mechanical wear and calibration drift --- a robotic echo of the pace-of-life syndrome, in which a faster tempo accompanies faster ageing \cite{Reale2010}. A parameter $\tau_{\text{act}}$ scales the maximum velocity bound or the control-effort penalty.

\item \textbf{Aggressiveness} governs assertiveness in competition for shared resources --- charging stations, bandwidth, contested space --- concretely, how yielding is apportioned during path deconfliction \cite{vandenberg2011}: an assertive robot claims right-of-way while a deferential one shoulders more of the avoidance responsibility \cite{Guo2021,Buckman2019}. A parameter $\tau_{\text{agg}}$ modulates this priority, so that higher-$\tau_{\text{agg}}$ robots impose more of the burden on others.

\item \textbf{Sociability} governs the tendency to aggregate or disperse: balancing spatial coverage against the proximity needed for local sensing and consensus. In communication-constrained swarms it shapes network topology --- sociable `anchor' robots holding a connected backbone while others range to the periphery \cite{Tarapore2020,Webber2023}. A parameter $\tau_{\text{soc}}$ scales the attractive component of an artificial potential field \cite{khatib1986real}, or the target algebraic connectivity of the communication graph \cite{Olfati-Saber2007}.
\end{itemize}

The mapping is many-to-many rather than one-to-one: a single mission requirement may load onto several axes (survivability draws on both shyness--boldness and activity level), and a single axis may serve several requirements. Its specific control-theoretic realisation depends on the deployment context, as the next section develops.

\paragraph*{Controller-agnostic implementation}

This vocabulary is illustrative, not prescriptive: the framework is agnostic to the robot's control mechanism --- the sense--decide--act architecture through which behaviour is computed --- and applies wherever that behaviour can be modulated by a low-dimensional parameter. In behaviour-based or subsumption architectures \cite{Brooks1986}, it adjusts module activation thresholds; in multi-agent reinforcement learning, the temperament vector conditions the policy \cite{marl-book}, letting one network express different temperaments at inference; in planners built on large language or vision-language-action models \cite{pmlr-v202-driess23a,pmlr-v229-zitkovich23a}, it can be encoded in the system prompt or used to filter candidate plans. In each case the temperament parameter operates above the controller's internal mechanism:  it does not replace the controller but shapes the distribution of behaviours it produces. The top-down phase supplies a prior, not a fixed target: it fixes the axes and a plausible distribution over them, narrowing the search enough that any bottom-up method --- such as evolutionary or learning-based, as swarm robotics commonly favours \cite{Ferrante2015} --- can search the remaining controller space efficiently. Such search sits naturally in the verification loop (Figure~\ref{fig3}, Panel 2), seeded by the top-down proposal but free to settle on a distribution the prior did not anticipate. This is the off-line design programme for swarms, in which behaviour is optimised in simulation before deployment \cite{Birattari2019,GarzonRamos2025}; there, expert-supplied template solutions can seed the search rather than be rediscovered each time --- the role played here by the top-down prior.

This controller agnosticism is distinct from the question of how coordination authority is distributed across the collective: temperament engineering's functional payoff is greatest under decentralisation, where a temperament distribution's anticipatory adaptation to the mission ecology substitutes for the global adaptivity a central planner would otherwise supply. A system that switches between centralised and self-organised coordination (e.g., \cite{zhu2024sons}) would accordingly carry a latent temperament distribution: masked while the planner dictates behaviour, and --- in its software-resident part --- reassigned or released when control is shed.

\subparagraph{Explainability and human oversight} 

This level of abstraction has direct consequences for explainability and human–robot interaction. The internal workings of a deep policy network or a foundation model are typically opaque to operators, but a temperament vector is not: a single human-readable parameter such as ``boldness = 0.2'' specifies in advance how the robot will weigh risk against reward across the decisions it faces. An operator who cannot reason about the latent dynamics of a learned controller can nonetheless reason about --- and adjust --- its temperament, and can predict the qualitative shape of its behaviour from that adjustment alone. Given that perceived temperament has been shown to influence human trust and collaborative efficiency in HRI \cite{Belgiovine2022,WilsonSmall2023,Tapus2008}, this offers a principled bridge between increasingly capable but opaque control mechanisms and the human supervisors who must monitor and steer them.

\subsection*{Shaping the distribution}

\paragraph{Distribution Planning} Once the relevant axes have been identified (Trait Mapping) and operationalised (above), and a control architecture chosen, Distribution Planning sketches out the shape of the $\tau$ distribution across the swarm on each axis. The behavioural hypervolume (the region a swarm occupies in the multi-dimensional space of temperament axes) gives a quantitative handle: a swarm spanning a larger region may be more functionally diverse and resilient. But the design lever is not only how much variation to introduce but what kind: Twu et al.~\cite{Twu2014} distinguish disparity (how different agents are, captured by the hypervolume) from complexity (how evenly they are spread). The two come apart --- an engineered bimodal distribution is high-disparity but low-complexity, spanning a wide range at two clustered points (Figure~\ref{fig1}c), whereas incidental temperament shows the inverse (Figure~\ref{fig1}b) --- and Distribution Planning chooses the kind of heterogeneity deliberately. As the field matures, recurring distributional motifs --- a bimodal split of bold and cautious robots, say --- may become part of a designer's repertoire.

\paragraph*{Plasticity via behavioural reaction norms}

The behavioural reaction norm (BRN) formalism from behavioural ecology gives a precise vocabulary for plasticity \cite{Dingemanse2010,Hunt2020plasticity,piersma2003phenotypic}. It decomposes behaviour into an elevation --- the baseline trait value set during Trait Mapping and Distribution Planning --- and a slope, the rate at which the trait changes along an environmental gradient such as battery voltage, local density, perceived threat or calibration drift. Plasticity Tuning specifies the slope: a robot's instantaneous parameter is $\tau(c) = \tau_0 + \beta(c - \bar{c})$, with $\tau_0$ the elevation, $\beta$ the slope, and $c$ a cue centred on its expected value. The linear form is only first-order --- $\beta$ is the local slope of a more general $\tau(c) = g(c)$, which need not be linear, as nonlinear reaction norms are common in nature \cite{Crowther2024}.

Plasticity tuning spans a static and a dynamic regime, and the dynamic regime admits two degrees. In the \textit{static} regime ($\beta = 0$) the robot holds its assigned $\tau$ whatever it meets. Holding $\tau$ fixed is a legitimate design choice, not merely the absence of plasticity. Even where plasticity is available, the optimal slope is rarely maximal: it is bounded by the cost of plasticity itself \cite{Haaland2021,Morgan2022}, and by the reliability of the cue $c$ as a predictor of the conditions the agent must match --- the less reliable the cue, the shallower the optimal reaction norm \cite{Bonamour2019}, and an over-steep response in a rapidly varying or noisy environment can lag or overshoot, actively degrading performance \cite{Reed2010}. A fixed temperament can therefore be efficient where the environment varies little, changes faster than the swarm can sense and respond, or offers no dependable cue to act on.

In the \textit{dynamic} regime ($\beta \neq 0$) temperament shifts with circumstance: in the first degree the reaction norm is fixed and the robot moves along it under sensor feedback --- scaling activity down, say, as battery voltage drops, to conserve the quality of its contribution; in the second, the robot also revises the norm itself, updating elevation and slope from accumulated experience. This last is a form of meta-learning (e.g., \cite{Richards2021}) --- a robot that not only adapts, but adapts how it adapts. 

These degrees have precedents in the narrower setting of task-allocation thresholds, which the reaction-norm formalism subsumes as one low-level realisation: Wu and Mathias show that a heterogeneous range of fixed thresholds enables specialisation while avoiding maladaptive sink states \cite{wu2020}, and Kazakova et al.\ that swarms respecialise by forgetting reinforced thresholds when demand shifts \cite{Kazakova2020} --- a threshold analogue of revising the reaction norm itself. The sim-to-real gap \cite{Tobin2017,Zhao2020,Birattari2019} is a natural case for employing the meta-learning regime: on detecting that its dynamics model mismatches real feedback, a robot can default to a cautious, low-activity state --- first-degree plasticity preserving its integrity --- while it revises the reaction norm itself from the accumulating mismatch evidence. Here the mismatch concerns the robot's own model rather than its environment, foreshadowing the self-model plasticity axis of the Outlook.

\subsection*{Spatial expression and verification}
A temperament distribution does not sit statically over the swarm: it expresses itself through movement. Activity governs how far a robot ranges, sociability whether it aggregates or disperses, boldness and exploration which regions it enters --- so the
distribution planned above induces a spatial distribution of robots, and with it the interaction structure determining who communicates and cooperates with whom. This is the link movement ecology draws between behavioural type and the spatial and
social structure a population generates \cite{Webber2023,Wolf2014,Croft2009}; and because emergent capability is a product of that structure, not the temperament distribution alone, the same composition
deployed into different environmental geometries \cite{Gordon2014,Pinter-Wollman2018} need not perform alike. The swarm's spatial expression is therefore emergent and resists accurate top-down prediction, which makes it the proper object of Phase~2 (Figure~\ref{fig3}): the verification scenarios, escalating to whole-mission rehearsal, simulate the realised spatial behaviour and confirm that it supports the intended functional outcomes before deployment.

\section*{From Individual Temperament to Collective Behaviour}

The five axes the framework rests on are not an arbitrary choice but are drawn from a mature body of behavioural-ecology theory. Since the early 2000s, the study of consistent individual differences in animals has grown rapidly under the labels `temperament', `personality' and `behavioural syndromes' \cite{Reale2007,Sih2004a,Sih2004b}, and such variation, once treated as statistical noise around an adaptive mean, is now recognised as ubiquitous, heritable and ecologically consequential \cite{Wolf2012}. The foundational framework of R\'eale et al.\ \cite{Reale2007} distinguishes five axes (Figure~\ref{fig2}), capturing fundamental trade-offs: safety versus opportunity, novelty versus familiarity, energy expenditure versus conservation, individual advantage versus group cohesion \cite{Sih2004a}.

These traits have measurable fitness consequences maintained by ecological trade-offs: bolder fish suffer higher avian predation \cite{Hulthen2017}, bolder seabirds expand their foraging range and raise offspring growth \cite{Pereira2024}, and a meta-analysis finds boldness buys reproductive success at a survival cost \cite{Smith2008}, while more broadly such differences shape population persistence and community dynamics \cite{Wolf2012}. They often covary as `behavioural syndromes' --- bold individuals also tending to be more aggressive and exploratory \cite{Sih2004a,Sih2004b} --- but where biological syndromes arise from genetic linkage and hormonal constraint, engineers can decouple the axes deliberately, abstracting the advantages of natural variation while engineering out its maladaptive correlations.

At the collective level, a group's temperament distribution matters not additively but emergently. A few highly exploratory fish can dictate the foraging routes of a whole shoal, heterogeneous groups often outperforming homogeneous ones \cite{Brown2014,Ioannou2017}; honeybee colonies hold a stable mix of fast-inaccurate and slow-accurate foragers as a bet-hedge raising nectar intake across flower qualities \cite{BurnsDyer2008}. These illustrate `collective personality', an emergent group-level phenotype not reducible to the average of individual ones \cite{Wray2011,Bengston2014a,Bengston2014}. The engineering counterpart is already visible: \citet{Kengyel2015} find an evolved behaviourally heterogeneous swarm outperforming every homogeneous baseline on aggregation, with the best mixture distinctly non-trivial; on the exploration--avoidance axis, simulated multi-robot patrols show a negatively skewed distribution --- mostly attentive robots, a single exploratory one --- outperforming homogeneous swarms once individuals share what they find \cite{York2024}. \citet{Piro2026} find mixed exploratory--exploitative swarms winning in turbulent olfactory search, because a fixed distribution of types resists spatial signal correlations that defeat an individually-balanced strategy. These remain scattered, largely single-axis results; assembling them into a systematic, multi-axis account is what temperament engineering sets out to do.

The behavioural hypervolume gives a quantitative handle on this collective diversity: adapted from the $n$-dimensional niche concept in ecology \cite{Hutchinson1957,Blonder2018,Takola2022}, it has been used to quantify the multi-dimensional behavioural-trait diversity of animal populations \cite{BastilleRousseau2019}, including the functional diversity of personality traits among conspecifics \cite{Mortelliti2020}. A swarm differs in that its variation is organised toward shared tasks, so capability is genuinely emergent: the value of any one behavioural type depends on the whole collective, and no single-robot controller optimisation can substitute for getting the distribution right --- yet the mature, quantitative vocabulary behavioural ecology has built for this variation is one swarm robotics has yet to adopt, even as it grapples with how to compare collective behaviours \cite{Jesus2026}.

\section*{Beyond Roles and Thresholds}

The framework set out above takes up a recognised challenge, posed but not yet met: operationalising swarm heterogeneity, in both hardware and control, as something the designer specifies \cite{Yang2018,Dorigo2020}. A survey of the landscape reveals a conspicuous gap: recent reviews of bio-inspired swarm robotics, while systematically mapping collective behaviour onto coordination and cooperation problems, treat variation among individuals as a matter of emergent dynamics or optimiser tuning rather than something the designer specifies \cite{Duan2023,Jiang2026}. Research on multi-robot heterogeneity has focused predominantly on morphological and functional role differentiation, or on algorithmic task allocation where robots are assigned to tasks based on capability or cost \cite{Rizk2019}. When behavioural variation is introduced, it typically takes the form of pre-specified roles or reactive threshold rules, without a unifying vocabulary for what kinds of behavioural variation matter and why. Emergent task specialisation in homogeneous groups of robots \cite{Ferrante2015} or response-threshold adaptation \cite{Bonabeau1997,Castello2016} produces behavioural variation as a byproduct of task allocation, rather than as a designed property addressing ecological trade-offs. Temperament parameters operate at a level above task-acceptance thresholds and can coexist with them rather than replacing them.

This gap matters because many of the hardest problems of field deployment are, in the end, behavioural ones. Robots must trade known risk against potential reward, balance exploiting known information against exploring the unknown, calibrate exertion against its energetic and mechanical costs, and choose whether to cluster or spread out. These decisions define each robot's behavioural strategy, and their distribution across the swarm shapes collective performance in ways no low-level controller optimisation can replace. The case for behavioural diversity of this kind has been made at the level of research programme by \citet{Ayanian2019}, whose `Diversity-enhanced Autonomy in Robot Teams' (DART) paradigm argues that a team should carry a complementary ensemble of control policies rather than a single policy optimised under one set of assumptions and deployed to every robot. Temperament engineering operationalises that programme, but relocates the diversity: it specifies which behavioural dimensions to vary, drawn from a validated vocabulary, and how to distribute them across the team, and it does so at the level of the temperament parameter $\tau$ rather than the policy, so that a swarm running one shared controller --- or one shared learned policy conditioned on $\tau$ --- can nonetheless carry an engineered behavioural distribution. Prorok et al.\ \cite{Prorok2017} considered heterogeneity of binary capability traits, drawing on the biodiversity literature \cite{Petchey2002} to specify capability distributions across the swarm's tasks. I instead consider continuous, temperament-axis variation across the swarm, anticipating its `mission ecology' in deployment rather than its task allocation --- a behavioural-strategy framework that complements such capability-heterogeneity work. Constraint-based robot ecology offers a further contrast: Egerstedt and colleagues argue that survivability constraints shape robot behaviour over long deployments \cite{Egerstedt2018,egerstedt2021robot}; temperament engineering shares that ecological intuition but operates at a different level, because constraint-based design optimises a single robot against its habitat whereas temperament engineering targets the emergent capability that arises when a distribution of strategies interacts among robots and their environment via self-organisation \cite{Camazine2003}.

\section*{Illustrative Deployment Scenarios}

\paragraph{Hazardous infrastructure inspection}

In nuclear decommissioning, robots perform remote inspection in constrained, highly
regulated environments \cite{NuclearDecom2025}. The shy--bold axis maps to the
probability of entering elevated-radiation zones. A `shy' temperament can be
guaranteed by a tight control barrier function (CBF, \cite{Ames2017}) that physically
bars the robot from zones above a radiation threshold, whatever its higher-level
goals. A `bold' scout runs a relaxed CBF, accepting higher known radiation than the shy threshold permits. The biological pattern of risk-taking minorities backed by a resilient cautious majority \cite{Brown2014,Ioannou2017} suggests bimodal boldness
distributions here.

\paragraph{Long-term environmental monitoring}

Sparse swarms operate over kilometre-scale environments where intermittent
connectivity and energy scarcity, not acute hazard, dominate \cite{Tarapore2020}. The
mission ecology rewards a broad behavioural hypervolume held stable over long
horizons. $\tau_{\text{exp}}$ governs how strongly each robot weights information gain against return-to-contact: low-$\tau_{\text{exp}}$ robots are persistent
path-followers keeping regular rendezvous with base stations, high-$\tau_{\text{exp}}$ outriders bias toward unvisited regions --- a coverage-versus-discovery bet-hedge like the honeybee forager mix \cite{BurnsDyer2008}. A closed-loop reaction norm scaling $\tau_{\text{exp}}$ down as battery voltage falls lets an outrider retreat to conservative path-following before it strands itself.

\paragraph{Assisted search-and-rescue}

In urban search-and-rescue, swarms can rapidly map interiors and assess structural
hazards under acute time pressure \cite{McGuire2019,Penders2010}. Unlike monitoring,
this rewards a deliberately skewed distribution: temperament engineering gives a
principled way to choose how many high-activity, high-boldness robots to commit under
constraints on battery and collapse risk. A small high-$\tau_{\text{bold}}$,
high-$\tau_{\text{act}}$ minority pushes into structurally uncertain volumes to find
survivors fast, while a cautious majority consolidates maps and holds relays. Because
the threat map is built online and unreliable early on, this is exactly the case for the meta-learning regime above: robots update their reaction norms as evidence of instability accumulates.

\section*{Outlook}

Temperament engineering treats the temperament distribution --- not any single controller --- as the design object, and the swarm's emergent collective capability as the design target. The most consequential work, I suspect, will lie not in any single axis but in the structure of the temperament space itself. That space is not a set of independent `dials': in nature the traits covary as syndromes \cite{Sih2004a}, and in swarms, engineered interactions between boldness, exploration, activity and sociability may instead produce nonlinear performance effects, tunable to a mission rather than inherited as evolutionary legacy. Prorok et al.\ \cite{Prorok2017} showed that as trait-space diversity grows, the set of distributions realising a given target shrinks, so that optimal configurations become harder to find; the same should hold along behavioural axes, where richer spaces carry a design cost of their own, and near-optimal rather than optimal $\tau$ distributions may prove the practical target --- a compromise swarm roboticists know well. This question --- when, and how much, heterogeneity repays its cost --- is beginning to receive formal treatment: within multi-agent reinforcement learning, \citet{amir2026} characterise the conditions under which behavioural diversity raises collective return, in broad cases reducing it to a tractable test on the reward structure. Temperament engineering poses the same problem one level up, axis by axis, where the design object is a distribution over validated behavioural trade-offs rather than a space of learned policies.

A further question is whether the axis set should stop at five (Figure~\ref{fig2}). The five are retained here as a validated core: each names a trade-off that natural selection has faced and negotiated, which is what lends the vocabulary its authority. A robot inherits all of them --- the biological axes sit wholly inside the robotic case. But where the animal is solely an evolved body, the robot is also designed and modelled, disposable rather than self-maintaining, and answerable to a human operator. Each of these is a respect in which the robot extends beyond the animal, and each raises a trade-off the five do not reach.

Because a robot's body is designed, and explicitly modelled, its self-model may be wrong, and may be revised: this invites a \textbf{`Self-Model Plasticity'} axis, a readiness to revise the internal dynamics-model \cite{Hoffmann2010,Chen2022} in the light of fresh sensorimotor evidence rather than hold to the prior --- where exploration--avoidance concerns novelty in the world, this concerns novelty in the self. Because its body is not self-maintaining, and can be made to yield or reconfigure, how far it resists or gives way on contact becomes a behavioural setting in its own right: a \textbf{`Forcefulness'} axis, trading precision and force transmission against energy absorption and safety on contact \cite{AbuDakka2020} --- distinct from aggressiveness in governing physical engagement rather than resource competition. And because it acts within a human-authored command structure, two further dimensions arise. \textbf{`Initiative'} fixes a robot's position on the authority gradient \cite{Beer2014}, how far it acts on its own judgement rather than deferring to the operator: a gradient evolution never had to negotiate, and one distinct from the competition among peers that aggressiveness captures. \textbf{`Expressiveness'} is the degree to which a robot shapes its behaviour to express its intent and remain legible to teammates and overseers \cite{Pascher2023}, trading raw efficiency for inferability. The two are largely orthogonal --- a robot may be deferential yet opaque, or high in initiative yet self-explaining --- but they interact, because a supervisor can only sensibly grant autonomy to a robot whose behaviour it can predict; and perceived temperament is already known to influence operator trust \cite{Belgiovine2022,WilsonSmall2023,Tapus2008}. Charting which of these dimensions are new, which extend a biological precedent, and which collide with one another is itself a programme worth pursuing.

Beyond enumerating axes, the framework's value is that it makes calibration a well-posed problem. Because it operates above the controller, it is indifferent to the controller's internals: low-level temperament parameters may override the unsafe plans of an otherwise-opaque foundation-model planner \cite{MonWilliams2025}, acting as a kind of embodied safety filter. It identifies which behavioural dimensions warrant variation and how to distribute them, without yet prescribing where on each axis the $\tau$ distribution should sit or how reaction norms should be set. It turns that calibration from an ad hoc choice into a problem with a finite vocabulary of axes, a formalism for distributions and plasticity, and a clear design object. A first foothold already exists --- engineered heterogeneity outperforming homogeneous swarms, in principle \cite{Kengyel2015} and on the exploration axis in dynamic conditions \cite{York2024,Piro2026}, for example --- but the case across the remaining axes, and across their interactions, is still to be made. Because swarm capability is emergent, the conditions under which an engineered distribution outperforms both homogeneous swarms and incidental variation \cite{Raoufi2023} must be established across the behavioural hypervolume, and against the real-world mission ecology, rather than axis by axis: the work the field can now take forward.

\backmatter

\section*{Declarations}

\begin{itemize}
\item Funding: E.R.H.\ is supported by the Royal Academy of Engineering under the Research Fellowship programme.  
\item Competing interests: The author declares no competing interests.
\item Ethics approval and consent to participate: Not applicable. 
\item Consent for publication: The author gives consent for publication.
\item Data and code availability:  Not applicable.
\item Author contribution: E.R.H.\ conceived the original concept, developed the theoretical framework, and wrote the manuscript. 
\end{itemize}

\bibliography{references_TE.bib}

@article{Reale2007,
  author  = {R{\'e}ale, Denis and Reader, Simon M. and Sol, Daniel and McDougall, Peter T. and Dingemanse, Niels J.},
  title   = {Integrating animal temperament within ecology and evolution},
  journal = {Biological Reviews},
  year    = {2007},
  volume  = {82},
  number  = {2},
  pages   = {291--318},
  doi     = {10.1111/j.1469-185X.2007.00010.x}
}

@article{Dingemanse2010,
  author  = {Dingemanse, Niels J. and Kazem, Anahita J. N. and R{\'e}ale, Denis and Wright, Jonathan},
  title   = {Behavioural reaction norms: animal personality meets individual plasticity},
  journal = {Trends in Ecology \& Evolution},
  year    = {2010},
  volume  = {25},
  number  = {2},
  pages   = {81--89},
  doi     = {10.1016/j.tree.2009.07.013}
}

@article{Ducatelle2011,
  author  = {Ducatelle, Frederick and Di Caro, Gianni A. and Pinciroli, Carlo and Gambardella, Luca M.},
  title   = {Self-organized cooperation between robotic swarms},
  journal = {Swarm Intelligence},
  year    = {2011},
  volume  = {5},
  number  = {2},
  pages   = {73--96},
  doi     = {10.1007/s11721-011-0053-0}
}

@incollection{Thenius2016,
  author    = {Thenius, Ronald and Moser, Daniel and Varughese, Joshua Cherian and Kernbach, Serge and Kuksin, Igor and Kernbach, Olga and Kuksina, Elena and Mi{\v{s}}kovi{\'c}, Nikola and Bogdan, Stjepan and Petrovi{\'c}, Tamara and Babi{\'c}, Anja and Boyer, Fr{\'e}d{\'e}ric and Lebastard, Vincent and Bazeille, St{\'e}phane and Ferrari, Graziano and Donati, Elisa and Pelliccia, Riccardo and Romano, Donato and Jansen van Vuuren, Godfried and Stefanini, Cesare and Morgantin, Matteo and Campo, Alexandre and Schmickl, Thomas},
  title     = {SubCULTron -- Cultural Development as a Tool in Underwater Robotics},
  booktitle = {Artificial Life and Intelligent Agents},
  publisher = {Springer},
  year      = {2018},
  pages     = {141--154},
  doi       = {10.1007/978-3-319-90418-4_3},
  address = {Cham}
}

@article{Prorok2017,
   author = {Amanda Prorok and M Ani Hsieh and Vijay Kumar},
   doi = {10.1109/TRO.2016.2631593},
   issue = {2},
   journal = {IEEE Transactions on Robotics},
   pages = {346-358},
   title = {The Impact of Diversity on Optimal Control Policies for Heterogeneous Robot Swarms},
   volume = {33},
   year = {2017}
}

@article{Petchey2002,
author = {Petchey, Owen L. and Gaston, Kevin J. },
title = {Functional diversity ({FD}), species richness and community composition},
journal = {Ecology Letters},
volume = {5},
number = {3},
pages = {402-411},
doi = {https://doi.org/10.1046/j.1461-0248.2002.00339.x},
year = {2002}
}

@article{Castello2016,
   author = {Eduardo Castello and Tomoyuki Yamamoto and Fabio Dalla Libera and Wenguo Liu and Alan F.T. Winfield and Yutaka Nakamura and Hiroshi Ishiguro},
   doi = {10.1007/s11721-015-0117-7},
   isbn = {8180909832},
   issn = {19353820},
   issue = {1},
   journal = {Swarm Intelligence},
   pages = {1-31},
   publisher = {Springer US},
   title = {Adaptive foraging for simulated and real robotic swarms: the dynamical response threshold approach},
   volume = {10},
   year = {2016}
}

@incollection{Bonabeau1997,
author = {Bonabeau, Eric and Deneubourg, Jean-Louis and Sobkowski, Andrej and Theraulaz, Guy},
editors = {Dan Lundh and Björn Olsson and Ajit Narayanan},
title = {Adaptive Task Allocation Inspired by a Model of Division of Labor in Social Insects},
booktitle = {Biocomputing and Emergent Computation -- Proceedings of BCEC97},
address = {Singapore},
pages = {36-45},
doi = {10.1142/9789814529242},
    publisher = {World Scientific Press},
    year = {1997}
}

@article{Ferrante2015,
   author = {Eliseo Ferrante and Ali Emre Turgut and Edgar Duéñez-Guzmán and Marco Dorigo and Tom Wenseleers},
   doi = {10.1371/journal.pcbi.1004273},
   issue = {8},
   journal = {PLOS Computational Biology},
   month = {8},
   pages = {e1004273},
   publisher = {Public Library of Science},
   title = {Evolution of Self-Organized Task Specialization in Robot Swarms},
   volume = {11},
   year = {2015}
}

@article{Jiang2026,
   author = {Andong Jiang and Youzhi Xu and Haibing Zhang and Xuefei Liu and Shikun Wen and Yi Sun and Ping Zhang and Qingfei Han and Aihong Ji},
   doi = {10.1088/1748-3190/ae5e33},
   issn = {17483190},
   issue = {3},
   journal = {Bioinspiration and Biomimetics},
   month = {5},
   publisher = {Institute of Physics},
   title = {From nature to robotics: insights of animals collective behaviors on the development of swarm intelligence and multi-robot systems},
   volume = {21},
   year = {2026}
}

@article{Duan2023,
    author = {Duan, Haibin and Huo, Mengzhen and Fan, Yanming},
    title = {From animal collective behaviors to swarm robotic cooperation},
    journal = {National Science Review},
    volume = {10},
    number = {5},
    pages = {nwad040},
    year = {2023},
    month = {05},
    issn = {2095-5138},
    doi = {10.1093/nsr/nwad040},
}

@article{Reale2010,
    author = {Réale, Denis and Garant, Dany and Humphries, Murray M. and Bergeron, Patrick and Careau, Vincent and Montiglio, Pierre-Olivier},
    title = {Personality and the emergence of the pace-of-life syndrome concept at the population level},
    journal = {Philosophical Transactions of the Royal Society B: Biological Sciences},
    volume = {365},
    number = {1560},
    pages = {4051-4063},
    year = {2010},
    month = {12},
    issn = {0962-8436},
    doi = {10.1098/rstb.2010.0208},
}

@article{Sih2004a,
  author  = {Sih, Andrew and Bell, Alison M. and Johnson, J. Chadwick},
  title   = {Behavioral syndromes: an ecological and evolutionary overview},
  journal = {Trends in Ecology \& Evolution},
  year    = {2004},
  volume  = {19},
  number  = {7},
  pages   = {372--378},
  doi     = {10.1016/j.tree.2004.04.009}
}

@article{Sih2004b,
  author  = {Sih, Andrew and Bell, Alison and Johnson, J. Chadwick and Ziemba, Robert E.},
  title   = {Behavioral syndromes: an integrative overview},
  journal = {The Quarterly Review of Biology},
  year    = {2004},
  volume  = {79},
  number  = {3},
  pages   = {241--277},
  doi     = {10.1086/422893}
}

@article{Smith2008,
  author  = {Smith, Brian R. and Blumstein, Daniel T.},
  title   = {Fitness consequences of personality: a meta-analysis},
  journal = {Behavioral Ecology},
  year    = {2008},
  volume  = {19},
  number  = {2},
  pages   = {448--455},
  doi     = {10.1093/beheco/arm144}
}

@article{Wolf2012,
  author  = {Wolf, Max and Weissing, Franz J.},
  title   = {Animal personalities: consequences for ecology and evolution},
  journal = {Trends in Ecology \& Evolution},
  year    = {2012},
  volume  = {27},
  number  = {8},
  pages   = {452--461},
  doi     = {10.1016/j.tree.2012.05.001}
}

@article{OShea-Wheller2021,
  author  = {O'Shea-Wheller, Thomas A. and Hunt, Edmund R. and Sasaki, Takao},
  title   = {Functional heterogeneity in superorganisms: emerging trends and concepts},
  journal = {Annals of the Entomological Society of America},
  year    = {2021},
  volume  = {114},
  number  = {5},
  pages   = {562--574},
  doi     = {10.1093/aesa/saaa039}
}

@article{Blonder2018,
  author  = {Blonder, Benjamin},
  title   = {Hypervolume concepts in niche- and trait-based ecology},
  journal = {Ecography},
  year    = {2018},
  volume  = {41},
  number  = {9},
  pages   = {1441--1455},
  doi     = {10.1111/ecog.03187}
}

@article{Dorigo2020,
  author  = {Dorigo, Marco and Theraulaz, Guy and Trianni, Vito},
  title   = {Reflections on the future of swarm robotics},
  journal = {Science Robotics},
  year    = {2020},
  volume  = {5},
  number  = {49},
  pages   = {eabe4385},
  doi     = {10.1126/scirobotics.abe4385}
}

@article{Rizk2019,
  author  = {Rizk, Yara and Awad, Mariette and Tunstel, Edward W.},
  title   = {Cooperative heterogeneous multi-robot systems: a survey},
  journal = {ACM Computing Surveys},
  year    = {2019},
  volume  = {52},
  number  = {2},
  pages   = {1--31},
  doi     = {10.1145/3303848}
}

@article{Tarapore2020,
  author  = {Tarapore, Danesh and Gro{\ss}, Roderich and Zauner, Klaus-Peter},
  title   = {Sparse robot swarms: moving swarms to real-world applications},
  journal = {Frontiers in Robotics and AI},
  year    = {2020},
  volume  = {7},
  pages   = {83},
  doi     = {10.3389/frobt.2020.00083}
}

@article{Hunt2020plasticity,
  author  = {Hunt, Edmund R.},
  title   = {Phenotypic plasticity provides a bioinspiration framework for minimal field swarm robotics},
  journal = {Frontiers in Robotics and AI},
  year    = {2020},
  volume  = {7},
  pages   = {23},
  doi     = {10.3389/frobt.2020.00023}
}

@article{Hoffmann2010,
  author  = {Hoffmann, Matej and Marques, Hugo and Arieta, Alejandro and Sumioka, Hidenobu and Lungarella, Max and Pfeifer, Rolf},
  title   = {Body Schema in Robotics: A Review},
  journal = {IEEE Transactions on Autonomous Mental Development},
  volume  = {2},
  number  = {4},
  pages   = {304--324},
  year    = {2010},
  doi     = {10.1109/TAMD.2010.2086454}
}

@InProceedings{Jesus2026,
doi = "10.1007/978-3-032-26123-6_32",
author="Jesus, Andr{\'e} Fialho
and Kuckling, Jonas",
editor="Gro{\ss}, Roderich
and Becker, Aaron T.
and Di Caro, Gianni A.
and Haghighat, Bahar
and Hsieh, M. Ani
and Abu-Aisheh, Razanne
and Talamali, Mohamed S.
and Dorigo, Marco",
title="How Swarms Differ: Challenges in Collective Behaviour Comparison",
booktitle="Swarm Intelligence",
year="2026",
publisher="Springer Nature Switzerland",
address="Cham",
pages="376--384",
isbn="978-3-032-26123-6"
}

@article{Mortelliti2020,
  author  = {Mortelliti, Alessio and Brehm, Allison M.},
  title   = {Environmental Heterogeneity and Population Density Affect the Functional Diversity of Personality Traits in Small Mammal Populations},
  journal = {Proceedings of the Royal Society B: Biological Sciences},
  volume  = {287},
  number  = {1940},
  pages   = {20201713},
  year    = {2020},
  doi     = {10.1098/rspb.2020.1713}
}

@article{GarzonRamos2025,
  author  = {Garz\'on Ramos, David and Pagnozzi, Federico and St\"utzle, Thomas and Birattari, Mauro},
  title   = {Automatic Design of Robot Swarms under Concurrent Design Criteria: A Study Based on Iterated F-Race},
  journal = {Advanced Intelligent Systems},
  volume  = {7},
  number  = {1},
  pages   = {2400332},
  year    = {2025},
  doi     = {10.1002/aisy.202400332}
}

@article{Chen2022,
  author  = {Chen, Boyuan and Kwiatkowski, Robert and Vondrick, Carl and Lipson, Hod},
  title   = {Full-Body Visual Self-Modeling of Robot Morphologies},
  journal = {Science Robotics},
  volume  = {7},
  number  = {68},
  pages   = {eabn1944},
  year    = {2022},
  doi     = {10.1126/scirobotics.abn1944}
}

@article{Reed2010,
    author = {Reed, Thomas E. and Waples, Robin S. and Schindler, Daniel E. and Hard, Jeffrey J. and Kinnison, Michael T.},
    title = {Phenotypic plasticity and population viability: the importance of environmental predictability},
    journal = {Proceedings of the Royal Society B: Biological Sciences},
    volume = {277},
    number = {1699},
    pages = {3391-3400},
    year = {2010},
    month = {06},
    issn = {0962-8452},
    doi = {10.1098/rspb.2010.0771},
    url = {https://doi.org/10.1098/rspb.2010.0771},
}

@article{Bonamour2019,
    author = {Bonamour, Suzanne and Chevin, Luis-Miguel and Charmantier, Anne and Teplitsky, Céline},
    title = {Phenotypic plasticity in response to climate change: the importance of cue variation},
    journal = {Philosophical Transactions of the Royal Society B: Biological Sciences},
    volume = {374},
    number = {1768},
    pages = {20180178},
    year = {2019},
    month = {01},
    issn = {0962-8436},
    doi = {10.1098/rstb.2018.0178},
    url = {https://doi.org/10.1098/rstb.2018.0178},
}

@article{Birattari2019,
  author  = {Birattari, Mauro and Ligot, Antoine and Bozhinoski, Darko and Brambilla, Manuele and Francesca, Gianpiero and Garattoni, Lorenzo and Garz\'on Ramos, David and Hasselmann, Ken and Kegeleirs, Miquel and Kuckling, Jonas and Pagnozzi, Federico and Roli, Andrea and Salman, Muhammad and St\"utzle, Thomas},
  title   = {Automatic Off-Line Design of Robot Swarms: A Manifesto},
  journal = {Frontiers in Robotics and AI},
  volume  = {6},
  pages   = {59},
  year    = {2019},
  doi     = {10.3389/frobt.2019.00059}
}

@article{AbuDakka2020,
  author  = {Abu-Dakka, Fares J. and Saveriano, Matteo},
  title   = {Variable Impedance Control and Learning---A Review},
  journal = {Frontiers in Robotics and AI},
  volume  = {7},
  pages   = {590681},
  year    = {2020},
  doi     = {10.3389/frobt.2020.590681}
}

@article{Beer2014,
  author  = {Beer, Jenay M. and Fisk, Arthur D. and Rogers, Wendy A.},
  title   = {Toward a Framework for Levels of Robot Autonomy in Human--Robot Interaction},
  journal = {Journal of Human-Robot Interaction},
  volume  = {3},
  number  = {2},
  pages   = {74--99},
  year    = {2014},
  doi     = {10.5898/JHRI.3.2.Beer}
}

@inproceedings{Pascher2023,
  author    = {Pascher, Max and Gruenefeld, Uwe and Schneegass, Stefan and Gerken, Jens},
  title     = {How to Communicate Robot Motion Intent: A Scoping Review},
  booktitle = {Proceedings of the 2023 CHI Conference on Human Factors in Computing Systems (CHI '23)},
  articleno = {409},
  pages     = {1--17},
  year      = {2023},
  publisher = {Association for Computing Machinery},
  address   = {New York, NY, USA},
  doi       = {10.1145/3544548.3580857}
}

@article{Yang2018,
  author  = {Yang, Guang-Zhong and Bellingham, Jim and Dupont, Pierre E. and Fischer, Peer and Floridi, Luciano and Full, Robert and Jacobstein, Neil and Kumar, Vijay and McNutt, Marcia and Merrifield, Robert and Nelson, Bradley J. and Scassellati, Brian and Taddeo, Mariarosaria and Taylor, Russell and Veloso, Manuela and Wang, Zhong Lin and Wood, Robert},
  title   = {The grand challenges of {Science Robotics}},
  journal = {Science Robotics},
  year    = {2018},
  volume  = {3},
  number  = {14},
  pages   = {eaar7650},
  doi     = {10.1126/scirobotics.aar7650}
}

@article{NuclearDecom2025,
  author  = {Mart{\'\i}nez-Mart{\'\i}n, Estefan{\'\i}a and others},
  title   = {From traditional robotic deployments towards assisted robotic deployments in nuclear decommissioning},
  journal = {Frontiers in Robotics and AI},
  year    = {2025},
  volume  = {12},
  pages   = {1432845},
  doi     = {10.3389/frobt.2025.1432845}
}

@article{Briffa2017,
    author = {Briffa, Mark},
    title = {Abandoning animal personality would cause obfuscation: a comment on {Beekman and Jordan}},
    journal = {Behavioral Ecology},
    volume = {28},
    number = {3},
    pages = {625-626},
    year = {2017},
    month = {05},
    issn = {1045-2249},
    doi = {10.1093/beheco/arx025},
}

@inproceedings{amir2026,
  title={When Is Diversity Rewarded in Cooperative Multi-Agent Learning?},
  author={Amir, Michael and Bettini, Matteo and Prorok, Amanda},
  year={2026},
  booktitle={International Conference on Learning Representations (ICLR)},
}

@article{Bell2017,
    author = {Bell, Alison M.},
    title = {There is no special sauce: a comment on {Beekman and Jordan}},
    journal = {Behavioral Ecology},
    volume = {28},
    number = {3},
    pages = {626-627},
    year = {2017},
    month = {05},
    issn = {1045-2249},
    doi = {10.1093/beheco/arx031},
}

@article{Dingemanse2017,
    author = {Dingemanse, Niels J.},
    title = {The role of personality research in contemporary behavioral ecology: a comment on {Beekman and Jordan}},
    journal = {Behavioral Ecology},
    volume = {28},
    number = {3},
    pages = {624-625},
    year = {2017},
    month = {05},
    issn = {1045-2249},
    doi = {10.1093/beheco/arx027},
}

@article{Beekman2017,
   author = {Madeleine Beekman and L. Alex Jordan},
   doi = {10.1093/beheco/arx022},
   issn = {14657279},
   issue = {3},
   journal = {Behavioral Ecology},
   month = {5},
   pages = {617-623},
   publisher = {Oxford University Press},
   title = {Does the field of animal personality provide any new insights for behavioral ecology?},
   volume = {28},
   year = {2017}
}

@inproceedings{Penders2010,
  author    = {Penders, Jacques and Alboul, Lyuba and Witkowski, Mark and Naghsh, Amir and Saez-Pons, Joan and Hernandez, Sebastian and Hailes, Stephen},
  title     = {A robot swarm assisting a human fire-fighter},
  booktitle = {Proceedings of the 2010 IEEE International Symposium on Safety, Security and Rescue Robotics},
  year      = {2010},
  pages     = {1--6},
  doi       = {10.1109/SSRR.2010.5981575}
}

@article{McGuire2019,
  author  = {McGuire, Kimberly N. and De Wagter, Christophe and Tuyls, Karl and Kappen, Hilbert J. and de Croon, Guido C. H. E.},
  title   = {Minimal navigation solution for a swarm of tiny flying robots to explore an unknown environment},
  journal = {Science Robotics},
  year    = {2019},
  volume  = {4},
  number  = {35},
  pages   = {eaaw9710},
  doi     = {10.1126/scirobotics.aaw9710}
}

@article{Hulthen2017,
  title={A predation cost to bold fish in the wild},
  author={Hulth{\'e}n, Kaj and Chapman, Ben B and Nilsson, P Anders and Hansson, Lars-Anders and Skov, Christian and Brodersen, Jakob and Vinterstare, Jerker and Br{\"o}nmark, Christer},
  journal={Scientific Reports},
  volume={7},
  number={1},
  pages={1239},
  year={2017},
  publisher={Nature Publishing Group},
  doi={10.1038/s41598-017-01270-w}
}

@article{Pereira2024,
  title={Boldness predicts foraging behaviour, habitat use and chick growth in a central place marine predator},
  author={Pereira, Jorge M and Ramos, Jaime A and Ceia, Filipe R and Kr{\"u}ger, Lucas and Marques, Ana M and Paiva, Vitor H},
  journal={Oecologia},
  volume={205},
  number={1},
  pages={135--147},
  year={2024},
  publisher={Springer},
  doi={10.1007/s00442-024-05557-4}
}

@inproceedings{Belgiovine2022,
  title={Towards an {HRI} Tutoring Framework for Long-term Personalization and Real-time Adaptation},
  author={Belgiovine, Giulia and Gonzalez-Billandon, Jonas and Sandini, Giulio and Rea, Francesco and Sciutti, Alessandra},
  booktitle={Adjunct Proceedings of the 30th ACM Conference on User Modeling, Adaptation and Personalization},
  pages={139--145},
  year={2022},
  doi={10.1145/3511047.3537689}
}

@inproceedings{WilsonSmall2023,
author = {Wilson-Small, Nialah Jenae and Goedicke, David and Petersen, Kirstin and Azenkot, Shiri},
title = {A Drone Teacher: Designing Physical Human-Drone Interactions for Movement Instruction},
year = {2023},
isbn = {9781450399647},
publisher = {Association for Computing Machinery},
address = {New York, NY, USA},
doi = {10.1145/3568162.3576985},
booktitle = {Proceedings of the 2023 ACM/IEEE International Conference on Human-Robot Interaction},
pages = {311–320},
numpages = {10},
location = {Stockholm, Sweden},
series = {HRI '23}
}

@article{MonWilliams2025,
  title={Embodied large language models enable robots to complete complex tasks in unpredictable environments},
  author={Mon-Williams, Ruaridh and Li, Gen and Long, Ran and Du, Wei and Lucas, Christopher G},
  journal={Nature Machine Intelligence},
  volume={7},
  number={4},
  pages={592--601},
  year={2025},
  publisher={Nature Publishing Group},
  doi={10.1038/s42256-025-01005-x}
}

@article{BastilleRousseau2019,
  title={Leveraging multidimensional heterogeneity in resource selection to define movement tactics of animals},
  author={Bastille-Rousseau, Guillaume and Wittemyer, George},
  journal={Ecology Letters},
  volume={22},
  number={9},
  pages={1417--1427},
  year={2019},
  publisher={Wiley Online Library},
  doi={10.1111/ele.13327}
}

@article{Brown2014,
  title={Individual personality traits influence group exploration in a feral guppy population},
  author={Brown, Culum and Irving, E.},
  journal={Behavioral Ecology},
  volume={25},
  number={1},
  pages={95--101},
  year={2014},
  publisher={Oxford University Press},
  doi={10.1093/beheco/art090}
}

@inproceedings{Richards2021,
  author       = {Richards, S. M. and Azizan, N. and Slotine, J.-J. E. and Pavone, M.},
  title        = {Adaptive-control-oriented meta-learning for nonlinear systems},
  booktitle    = {Robotics: Science and Systems},
  year         = {2021},
  doi          = {10.15607/RSS.2021.XVII.056},
}

@InProceedings{Kengyel2015,
author="Kengyel, Daniela
and Hamann, Heiko
and Zahadat, Payam
and Radspieler, Gerald
and Wotawa, Franz
and Schmickl, Thomas",
editor="Chen, Qingliang
and Torroni, Paolo
and Villata, Serena
and Hsu, Jane
and Omicini, Andrea",
title="Potential of Heterogeneity in Collective Behaviors: A Case Study on Heterogeneous Swarms",
booktitle="PRIMA 2015: Principles and Practice of Multi-Agent Systems",
year="2015",
publisher="Springer International Publishing",
address="Cham",
pages="201--217",
isbn="978-3-319-25524-8"
}

@article{Piro2026,
  title = {Policy heterogeneity improves collective olfactory search in three-dimensional turbulence},
  author = {Piro, Lorenzo and Heinonen, Robin A. and Carbone, Maurizio and Biferale, Luca and Cencini, Massimo},
  journal = {Phys. Rev. E},
  volume = {113},
  issue = {4},
  pages = {044401},
  numpages = {7},
  year = {2026},
  month = {Apr},
  publisher = {American Physical Society},
  doi = {10.1103/6zls-m67c},
  url = {https://link.aps.org/doi/10.1103/6zls-m67c}
}

@article{Ioannou2017,
  title={High-predation habitats affect the social dynamics of collective exploration in a shoaling fish},
  author={Ioannou, Christos C and Ramnarine, Indar W and Torney, Colin J},
  journal={Science Advances},
  volume={3},
  number={5},
  pages={e1602682},
  year={2017},
  publisher={American Association for the Advancement of Science},
  doi={10.1126/sciadv.1602682}
}

@inproceedings{Zakir2024,
  author    = {Zakir, Raina and Salahshour, Mohammad and Dorigo, Marco and Reina, Andreagiovanni},
  title     = {Heterogeneity Can Enhance the Adaptivity of Robot Swarms to Dynamic Environments},
  booktitle = {Swarm Intelligence: 14th International Conference, ANTS 2024},
  series    = {Lecture Notes in Computer Science},
  volume    = {14987},
  pages     = {112--126},
  publisher = {Springer},
  address = {Cham},
  year      = {2024},
  doi       = {10.1007/978-3-031-70932-6_9}
}

@inproceedings{York2024,
  author    = {York, Connor J. and Madin, Zachary R. and O'Dowd, Paul and Hunt, Edmund R.},
  title     = {Shaping Multi-Robot Patrol Performance with Heterogeneity in Individual Learning Behavior},
  booktitle = {2024 IEEE International Conference on Development and Learning (ICDL)},
  year      = {2024},
  publisher = {IEEE},
  address = {Piscataway, NJ},
  doi       = {10.1109/ICDL61372.2024.10645013}
}

@inproceedings{Raoufi2023,
    author = {Raoufi, Mohsen and Romanczuk, Pawel and Hamann, Heiko},
    title = {Individuality in Swarm Robots with the Case Study of Kilobots: Noise, Bug, or Feature?},

    booktitle = {ALIFE 2023: Ghost in the Machine: Proceedings of the 2023 Artificial Life Conference},
    pages = {35},
    year = {2023},
    month = {07},
    doi = {10.1162/isal\_a\_00623},
}

@InProceedings{pmlr-v202-driess23a,
  title = 	 {{P}a{LM}-E: An Embodied Multimodal Language Model},
  author =       {Driess, Danny and Xia, Fei and Sajjadi, Mehdi S. M. and Lynch, Corey and Chowdhery, Aakanksha and Ichter, Brian and Wahid, Ayzaan and Tompson, Jonathan and Vuong, Quan and Yu, Tianhe and Huang, Wenlong and Chebotar, Yevgen and Sermanet, Pierre and Duckworth, Daniel and Levine, Sergey and Vanhoucke, Vincent and Hausman, Karol and Toussaint, Marc and Greff, Klaus and Zeng, Andy and Mordatch, Igor and Florence, Pete},
  booktitle = 	 {Proceedings of the 40th International Conference on Machine Learning},
  address = {Cambridge, MA},
  pages = 	 {8469--8488},
  year = 	 {2023},
  editor = 	 {Krause, Andreas and Brunskill, Emma and Cho, Kyunghyun and Engelhardt, Barbara and Sabato, Sivan and Scarlett, Jonathan},
  volume = 	 {202},
  series = 	 {Proceedings of Machine Learning Research},
  month = 	 {23--29 Jul},
  publisher =    {PMLR},
  url = 	 {https://proceedings.mlr.press/v202/driess23a.html}
}

@article{Crowther2024,
    author = {Crowther, Claudia and Bonser, Stephen P and Schwanz, Lisa E},
    title = {Plasticity and the adaptive evolution of switchlike reaction norms under environmental change},
    journal = {Evolution Letters},
    volume = {8},
    number = {1},
    pages = {64-75},
    year = {2024},
    month = {02},
    issn = {2056-3744},
    doi = {10.1093/evlett/qrad035},
}

@ARTICLE{Guo2021,
  author={Guo, Ke and Wang, Dawei and Fan, Tingxiang and Pan, Jia},
  journal={IEEE Robotics and Automation Letters}, 
  title={VR-ORCA: Variable Responsibility Optimal Reciprocal Collision Avoidance}, 
  year={2021},
  volume={6},
  number={3},
  pages={4520-4527},
  doi={10.1109/LRA.2021.3067851}}

@InProceedings{vandenberg2011,
doi = "10.1007/978-3-642-19457-3_1",
author="{van den Berg}, Jur and Guy, Stephen J.
and Lin, Ming
and Manocha, Dinesh",
editor="Pradalier, C{\'e}dric
and Siegwart, Roland
and Hirzinger, Gerhard",
title="Reciprocal n-Body Collision Avoidance",
booktitle="Robotics Research",
year="2011",
publisher="Springer Berlin Heidelberg",
address="Berlin, Heidelberg",
pages="3--19",
isbn="978-3-642-19457-3"
}

@book{Camazine2003,
   author = {Scott Camazine and Jean-Louis Deneubourg and Nigel R Franks and James Sneyd and Guy Theraulaz and Eric Bonabeau},
   address = {Princeton, NJ},
   isbn = {0691116245},
   publisher = {Princeton University Press},
   title = {Self-Organization in Biological Systems},
   year = {2001}
}

@article{Julian2014,
author = {Brian J. Julian and Sertac Karaman and Daniela Rus},
title ={On mutual information-based control of range sensing robots for mapping applications},
journal = {The International Journal of Robotics Research},
volume = {33},
number = {10},
pages = {1375-1392},
year = {2014},
doi = {10.1177/0278364914526288},
}

@InProceedings{Majumdar2020,
doi = "10.1007/978-3-030-28619-4_10",
author="Majumdar, Anirudha
and Pavone, Marco",
editor="Amato, Nancy M.
and Hager, Greg
and Thomas, Shawna
and Torres-Torriti, Miguel",
title="How Should a Robot Assess Risk? Towards an Axiomatic Theory of Risk in Robotics",
booktitle="Robotics Research",
year="2020",
publisher="Springer International Publishing",
address="Cham",
pages="75--84",
isbn="978-3-030-28619-4"
}

@article{Ayanian2019,
   author = {Nora Ayanian},
   doi = {10.1177/0278364919839137},
   issue = {12-13},
   journal = {The International Journal of Robotics Research},
   pages = {1329-1337},
   title = {{DART}: Diversity-enhanced Autonomy in Robot Teams},
   volume = {38},
   year = {2019}
}

@article{piersma2003phenotypic,
title = {Phenotypic flexibility and the evolution of organismal design},
journal = {Trends in Ecology \& Evolution},
volume = {18},
number = {5},
pages = {228-233},
year = {2003},
issn = {0169-5347},
doi = {10.1016/S0169-5347(03)00036-3},
author = {Theunis Piersma and Jan Drent}
}

@INPROCEEDINGS{Buckman2019,
  author={Buckman, Noam and Pierson, Alyssa and Schwarting, Wilko and Karaman, Sertac and Rus, Daniela},
  booktitle={2019 IEEE/RSJ International Conference on Intelligent Robots and Systems (IROS)}, 
  title={Sharing is Caring: Socially-Compliant Autonomous Intersection Negotiation}, 
  year={2019},
  volume={},
  number={},
  pages={6136-6143},
  doi={10.1109/IROS40897.2019.8967997}}

@InProceedings{Melhuish2007,
author="Melhuish, Chris
and Kubo, Masao",
editor="Alami, Rachid
and Chatila, Raja
and Asama, Hajime",
title="Collective Energy Distribution: Maintaining the Energy Balance in Distributed Autonomous Robots using Trophallaxis",
booktitle="Distributed Autonomous Robotic Systems 6",
year="2007",
doi = "10.1007/978-4-431-35873-2_27",
publisher="Springer Japan",
address="Tokyo",
pages="275--284",
isbn="978-4-431-35873-2"
}

@article{Charbonneau2017,
    author = {Charbonneau, Daniel and Poff, Corey and Nguyen, Hoan and Shin, Min C. and Kierstead, Karen and Dornhaus, Anna},
    title = {Who Are the “Lazy” Ants? The Function of Inactivity in Social Insects and a Possible Role of Constraint: Inactive Ants Are Corpulent and May Be Young and/or Selfish},
    journal = {Integrative and Comparative Biology},
    volume = {57},
    number = {3},
    pages = {649-667},
    year = {2017},
    month = {09},
    issn = {1540-7063},
    doi = {10.1093/icb/icx029},
}

@article{Wray2011,
   author = {Margaret K Wray and Heather R Mattila and Thomas D Seeley},
   doi = {10.1016/j.anbehav.2010.11.027},
   isbn = {0003-3472},
   issue = {3},
   journal = {Animal Behaviour},
   pages = {559-568},
   title = {Collective personalities in honeybee colonies are linked to colony fitness},
   volume = {81},
   year = {2011}
}

@article{Bengston2014,
   author = {S E Bengston and A Dornhaus},
   doi = {10.1098/rspb.2014.0518},
   isbn = {0962-8452
1471-2954},
   issue = {1791},
   journal = {Proceedings of the Royal Society B: Biological Sciences},
   pages = {20140518},
   publisher = {The Royal Society},
   title = {Be meek or be bold? A colony-level behavioural syndrome in ants},
   volume = {281},
   year = {2014}
}

@article{Bengston2014a,
   author = {Sarah E Bengston and Jennifer M Jandt},
   doi = {10.3389/fevo.2014.00081},
   isbn = {2296-701X},
   issue = {81},
   journal = {Frontiers in Ecology and Evolution},
   title = {The development of collective personality: the ontogenetic drivers of behavioral variation across groups},
   volume = {2},
   url = {http://journal.frontiersin.org/article/10.3389/fevo.2014.00081},
   year = {2014}
}

@book{egerstedt2021robot,
  title={Robot ecology: constraint-based design for long-duration autonomy},
  author={Egerstedt, Magnus},
  year={2021},
  publisher={Princeton University Press},
  address = {Princeton, NJ}
}

@ARTICLE{Olfati-Saber2007,
  author={Olfati-Saber, Reza and Fax, J. Alex and Murray, Richard M.},
  journal={Proceedings of the IEEE}, 
  title={Consensus and Cooperation in Networked Multi-Agent Systems}, 
  year={2007},
  volume={95},
  number={1},
  pages={215-233},
  doi={10.1109/JPROC.2006.887293}}

@article{khatib1986real,
  title={Real-time obstacle avoidance for manipulators and mobile robots},
  author={Khatib, Oussama},
  journal={The international journal of robotics research},
  volume={5},
  number={1},
  doi = {10.1177/027836498600500106},
  pages={90--98},
  year={1986},
  publisher={Sage Publications Sage CA: Thousand Oaks, CA}
}

@InProceedings{pmlr-v229-zitkovich23a,
  title = 	 {RT-2: Vision-Language-Action Models Transfer Web Knowledge to Robotic Control},
  author =       {Zitkovich, Brianna and Yu, Tianhe and Xu, Sichun and Xu, Peng and Xiao, Ted and Xia, Fei and Wu, Jialin and Wohlhart, Paul and Welker, Stefan and Wahid, Ayzaan and Vuong, Quan and Vanhoucke, Vincent and Tran, Huong and Soricut, Radu and Singh, Anikait and Singh, Jaspiar and Sermanet, Pierre and Sanketi, Pannag R. and Salazar, Grecia and Ryoo, Michael S. and Reymann, Krista and Rao, Kanishka and Pertsch, Karl and Mordatch, Igor and Michalewski, Henryk and Lu, Yao and Levine, Sergey and Lee, Lisa and Lee, Tsang-Wei Edward and Leal, Isabel and Kuang, Yuheng and Kalashnikov, Dmitry and Julian, Ryan and Joshi, Nikhil J. and Irpan, Alex and Ichter, Brian and Hsu, Jasmine and Herzog, Alexander and Hausman, Karol and Gopalakrishnan, Keerthana and Fu, Chuyuan and Florence, Pete and Finn, Chelsea and Dubey, Kumar Avinava and Driess, Danny and Ding, Tianli and Choromanski, Krzysztof Marcin and Chen, Xi and Chebotar, Yevgen and Carbajal, Justice and Brown, Noah and Brohan, Anthony and Arenas, Montserrat Gonzalez and Han, Kehang},
  booktitle = 	 {Proceedings of The 7th Conference on Robot Learning},
  address = {Cambridge, MA},
  pages = 	 {2165--2183},
  year = 	 {2023},
  editor = 	 {Tan, Jie and Toussaint, Marc and Darvish, Kourosh},
  volume = 	 {229},
  series = 	 {Proceedings of Machine Learning Research},
  month = 	 {06--09 Nov},
  publisher =    {PMLR},
  url = 	 {https://proceedings.mlr.press/v229/zitkovich23a.html}
}

@book{marl-book,
  author = {Stefano V. Albrecht and Filippos Christianos and Lukas Sch\"afer},
  title = {Multi-Agent Reinforcement Learning: Foundations and Modern Approaches},
  publisher = {MIT Press},
  address = {Cambridge, MA},
  year = {2024},
  url = {https://www.marl-book.com}
}

@INPROCEEDINGS{Zhao2020,
  author={Zhao, Wenshuai and Queralta, Jorge Peña and Westerlund, Tomi},
  booktitle={2020 IEEE Symposium Series on Computational Intelligence (SSCI)}, 
  title={Sim-to-Real Transfer in Deep Reinforcement Learning for Robotics: a Survey}, 
  year={2020},
  volume={},
  number={},
  pages={737-744},
  doi={10.1109/SSCI47803.2020.9308468}}

@INPROCEEDINGS{Tobin2017,
  author={Tobin, Josh and Fong, Rachel and Ray, Alex and Schneider, Jonas and Zaremba, Wojciech and Abbeel, Pieter},
  booktitle={2017 IEEE/RSJ International Conference on Intelligent Robots and Systems (IROS)}, 
  title={Domain randomization for transferring deep neural networks from simulation to the real world}, 
  year={2017},
  volume={},
  number={},
  pages={23-30},
  doi={10.1109/IROS.2017.8202133}}

@book{Pfeifer2006,
    author = {Pfeifer, Rolf and Bongard, Josh},
    title = {How the Body Shapes the Way We Think: A New View of Intelligence},
    publisher = {The MIT Press},
    year = {2006},
    month = {10},
    address = {Cambridge, MA},
    isbn = {9780262281553},
    doi = {10.7551/mitpress/3585.001.0001},
}

@article{Rus2015,
   author = {Daniela Rus and Michael T Tolley},
   doi = {10.1038/nature14543},
   issn = {1476-4687},
   issue = {7553},
   journal = {Nature},
   pages = {467-475},
   title = {Design, fabrication and control of soft robots},
   volume = {521},
   url = {https://doi.org/10.1038/nature14543},
   year = {2015}
}

@ARTICLE{Brooks1986,
  author={Brooks, R.},
  journal={IEEE Journal on Robotics and Automation}, 
  title={A robust layered control system for a mobile robot}, 
  year={1986},
  volume={2},
  number={1},
  pages={14-23},
  doi={10.1109/JRA.1986.1087032}}

@ARTICLE{Ames2017,
  author={Ames, Aaron D. and Xu, Xiangru and Grizzle, Jessy W. and Tabuada, Paulo},
  journal={IEEE Transactions on Automatic Control}, 
  title={Control Barrier Function Based Quadratic Programs for Safety Critical Systems}, 
  year={2017},
  volume={62},
  number={8},
  pages={3861-3876},
  doi={10.1109/TAC.2016.2638961}}

@article{Egerstedt2018,
   author = {Magnus Egerstedt and Jonathan N Pauli and Gennaro Notomista and Seth Hutchinson},
   doi = {https://doi.org/10.1016/j.arcontrol.2018.09.006},
   issn = {1367-5788},
   journal = {Annual Reviews in Control},
   pages = {1-7},
   title = {Robot ecology: Constraint-based control design for long duration autonomy},
   volume = {46},
   year = {2018}
}

@article{Haaland2021,
    author = {Haaland, Thomas R. and Wright, Jonathan and Ratikainen, Irja I.},
    title = {Individual reversible plasticity as a genotype‐level bet‐hedging strategy},
    journal = {Journal of Evolutionary Biology},
    volume = {34},
    number = {7},
    pages = {1022-1033},
    year = {2021},
    month = {07},
    issn = {1010-061X},
    doi = {10.1111/jeb.13788},
    url = {https://doi.org/10.1111/jeb.13788},
}

@article{Morgan2022,
author = {Rachael Morgan  and Anna H. Andreassen  and Eirik R. Åsheim  and Mette H. Finnøen  and Gunnar Dresler  and Tore Brembu  and Adrian Loh  and Joanna J. Miest  and Fredrik Jutfelt },
title = {Reduced physiological plasticity in a fish adapted to stable temperatures},
journal = {Proceedings of the National Academy of Sciences},
volume = {119},
number = {22},
pages = {e2201919119},
year = {2022},
doi = {10.1073/pnas.2201919119},
URL = {https://www.pnas.org/doi/abs/10.1073/pnas.2201919119}}

@InProceedings{Beni2005,
author="Beni, Gerardo",
editor="{\c{S}}ahin, Erol
and Spears, William M.",
title="From Swarm Intelligence to Swarm Robotics",
booktitle="Swarm Robotics",
year="2005",
publisher="Springer Berlin Heidelberg",
address="Berlin, Heidelberg",
pages="1--9",
isbn="978-3-540-30552-1"
}

@book{hamann2018,
   author = {Heiko Hamann},
   doi = {10.1007/978-3-319-74528-2},
   publisher = {Springer International Publishing},
   address = {Cham},
   title = {Swarm Robotics: A Formal Approach},
   year = {2018}
}

@INPROCEEDINGS{Twu2014,
  author={Twu, Philip and Mostofi, Yasamin and Egerstedt, Magnus},
  booktitle={2014 American Control Conference}, 
  title={A measure of heterogeneity in multi-agent systems}, 
  year={2014},
  volume={},
  number={},
  pages={3972-3977},
  doi={10.1109/ACC.2014.6858632}}

@article{Tapus2008,
   author = {Adriana Tapus and Cristian Ţăpuş and Maja J Matarić},
   doi = {10.1007/s11370-008-0017-4},
   issn = {1861-2784},
   issue = {2},
   journal = {Intelligent Service Robotics},
   pages = {169-183},
   title = {User—robot personality matching and assistive robot behavior adaptation for post-stroke rehabilitation therapy},
   volume = {1},
   url = {https://doi.org/10.1007/s11370-008-0017-4},
   year = {2008}
}

@article{Takola2022,
   author = {Elina Takola and Holger Schielzeth},
   doi = {10.1007/s10539-022-09849-y},
   issn = {15728404},
   issue = {4},
   journal = {Biology and Philosophy},
   month = {8},
   publisher = {Springer Science and Business Media B.V.},
   title = {Hutchinson’s ecological niche for individuals},
   volume = {37},
   year = {2022}
}

@article{Hutchinson1957,
author = {Hutchinson, G. Evelyn}, 
title = {Concluding Remarks},
volume = {22}, 
pages = {415-427}, 
year = {1957}, 
doi = {10.1101/SQB.1957.022.01.039}, 
journal = {Cold Spring Harbor Symposia on Quantitative Biology} 
}

@article{GaonaGordillo2023,
author = {Gaona-Gordillo, Irene and Holtmann, Benedikt and Mouchet, Alexia and Hutfluss, Alexander and Sánchez-Tójar, Alfredo and Dingemanse, Niels J.},
title = {Are animal personality, body condition, physiology and structural size integrated? A comparison of species, populations and sexes, and the value of study replication},
journal = {Journal of Animal Ecology},
volume = {92},
number = {9},
pages = {1707-1718},
doi = {https://doi.org/10.1111/1365-2656.13966},
year = {2023}
}

@article{Kazakova2020,
   author = {Vera A Kazakova and Annie S Wu and Gita R Sukthankar},
   doi = {10.1007/s11721-020-00181-3},
   issn = {1935-3820},
   issue = {3},
   journal = {Swarm Intelligence},
   pages = {171-204},
   title = {Respecializing swarms by forgetting reinforced thresholds},
   volume = {14},
   url = {https://doi.org/10.1007/s11721-020-00181-3},
   year = {2020}
}

@InProceedings{wu2020,
author="Wu, Annie S.
and Mathias, H. David",
editor="Dorigo, Marco
and St{\"u}tzle, Thomas
and Blesa, Maria J.
and Blum, Christian
and Hamann, Heiko
and Heinrich, Mary Katherine
and Strobel, Volker",
title="Dynamic Response Thresholds: Heterogeneous Ranges Allow Specialization While Mitigating Convergence to Sink States",
booktitle="Swarm Intelligence",
year="2020",
publisher="Springer International Publishing",
address="Cham",
pages="107--120",
isbn="978-3-030-60376-2"
}

@article{Pinter-Wollman2018,
    author = {Pinter-Wollman, Noa and Penn, Alan and Theraulaz, Guy and Fiore, Stephen M.},
    title = {Interdisciplinary approaches for uncovering the impacts of architecture on collective behaviour},
    journal = {Philosophical Transactions of the Royal Society B: Biological Sciences},
    volume = {373},
    number = {1753},
    pages = {20170232},
    year = {2018},
    month = {07},
    issn = {0962-8436},
    doi = {10.1098/rstb.2017.0232},
    url = {https://doi.org/10.1098/rstb.2017.0232},
}

@article{Gordon2014,
    doi = {10.1371/journal.pbio.1001805},
    author = {Gordon, Deborah M.},
    journal = {PLOS Biology},
    publisher = {Public Library of Science},
    title = {The Ecology of Collective Behavior},
    year = {2014},
    month = {03},
    volume = {12},
    url = {https://doi.org/10.1371/journal.pbio.1001805},
    pages = {1-4},
    number = {3},

}

@article{Croft2009,
   author = {Darren P. Croft and Jens Krause and Safi K. Darden and Indar W. Ramnarine and Jolyon J. Faria and Richard James},
   doi = {10.1007/s00265-009-0802-x},
   issn = {03405443},
   issue = {10},
   journal = {Behavioral Ecology and Sociobiology},
   pages = {1495-1503},
   title = {Behavioural trait assortment in a social network: Patterns and implications},
   volume = {63},
   year = {2009}
}

@article{Wolf2014,
   author = {Max Wolf and Jens Krause},
   doi = {10.1016/j.tree.2014.03.008},
   issn = {01695347},
   issue = {6},
   journal = {Trends in Ecology and Evolution},
   pages = {306-308},
   pmid = {24679987},
   publisher = {Elsevier Ltd},
   title = {Why personality differences matter for social functioning and social structure},
   volume = {29},
   year = {2014}
}

@article{
zhu2024sons,
author = {Weixu Zhu  and Sinan Oğuz  and Mary Katherine Heinrich  and Michael Allwright  and Mostafa Wahby  and Anders Lyhne Christensen  and Emanuele Garone  and Marco Dorigo },
title = {Self-organizing nervous systems for robot swarms},
journal = {Science Robotics},
volume = {9},
number = {96},
pages = {eadl5161},
year = {2024},
doi = {10.1126/scirobotics.adl5161}}

@ARTICLE{Li2025EMAS,
  author={Li, Zhuo and Wu, Weiran and Guo, Yunlong and Sun, Jian and Han, Qing-Long},
  journal={IEEE/CAA Journal of Automatica Sinica}, 
  title={Embodied Multi-Agent Systems: A Review}, 
  year={2025},
  volume={12},
  number={6},
  pages={1095-1116},
  doi={10.1109/JAS.2025.125552}}

@article{BurnsDyer2008,
  author  = {Burns, James G. and Dyer, Adrian G.},
  title   = {Diversity of speed-accuracy strategies benefits social insects},
  journal = {Current Biology},
  year    = {2008},
  volume  = {18},
  number  = {20},
  pages   = {R953--R954},
  doi     = {10.1016/j.cub.2008.08.028}
}

@article{Webber2023,
  author  = {Webber, Quinn M. R. and Albery, Gregory F. and Farine, Damien R. and Pinter-Wollman, Noa and Sharma, Nitika and Spiegel, Orr and Vander Wal, Eric and Manlove, Kezia},
  title   = {Behavioural ecology at the spatial--social interface},
  journal = {Biological Reviews},
  year    = {2023},
  volume  = {98},
  number  = {3},
  pages   = {868--886},
  doi     = {10.1111/brv.12934}
}

\end{document}